\documentclass{article}
\usepackage{spconf,amsmath,graphicx,amssymb,caption,subcaption}

\def\x{{\mathbf x}}
\def\y{{\mathbf y}}
\def\R{{\mathbb R}}
\def\ll{{\mathcal L}}
\def\wer{{\mathcal W}}
\def\G{{\mathcal G}}
\def\h{{\mathbf{h}}}
\def\cv{{\mathbf{c}}}
\def\sos{{\left<\texttt{sos}\right>}}
\def\eos{{\left<\texttt{eos}\right>}}

\title{Minimum Word Error Rate Training for Attention-based Sequence-to-Sequence
Models}
\name{Rohit Prabhavalkar \qquad Tara N. Sainath \qquad Yonghui Wu \qquad Patrick Nguyen\thanks{The
      authors would like to thank Matt Shannon, Erik McDermott, Michiel Bacchiani and
      Ha\c{s}im Sak for helpful comments and suggestions on this work.}}
\nameplus{Zhifeng Chen \qquad Chung-Cheng Chiu \qquad Anjuli Kannan}
\address{Google Inc.\\
	{\small \tt \{prabhavalkar,tsainath,yonghui,drpng,zhifengc,chungchengc,anjuli\}@google.com}}
\begin{document}
\ninept
\maketitle
\begin{abstract}
  Sequence-to-sequence models, such as attention-based models in automatic
  speech recognition (ASR), are typically trained to optimize the cross-entropy
  criterion which corresponds to improving the log-likelihood of the data.
  However, system performance is usually measured in terms of word error rate
  (WER), not log-likelihood.
  Traditional ASR systems benefit from discriminative sequence training which
  optimizes criteria such as the state-level minimum Bayes risk (sMBR) which are
  more closely related to WER.

  In the present work, we explore techniques to train attention-based models to
  directly minimize expected word error rate.
  We consider two loss functions which approximate the expected number of word
  errors: either by sampling from the model, or by using N-best lists of decoded
  hypotheses, which we find to be more effective than the sampling-based method.
  In experimental evaluations, we find that the proposed training procedure
  improves performance by up to 8.2\% relative to the baseline system.
  This allows us to train grapheme-based, uni-directional attention-based models
  which match the performance of a traditional, state-of-the-art, discriminative
  sequence-trained system on a mobile voice-search task.
\end{abstract}
\section{Introduction}
\label{sec:intro}
There has been growing interest in the automatic speech recognition (ASR)
community in building end-to-end trained, sequence-to-sequence models which
directly output a word sequence given input speech frames, without requiring
explicit alignments between the speech frames and labels.
Examples of such approaches include the recurrent neural network transducer
(RNN-T)~\cite{Graves12, GravesMohamedHinton13}, the recurrent neural aligner
(RNA)~\cite{SakShannonRaoEtAl17}, attention-based
models~\cite{ChanJaitlyLeEtAl15, BahdanauChorowskiSerdyukEtAl15}, and
connectionist temporal classification (CTC)~\cite{GravesFernandezGomezEtAl06}
with word-based targets~\cite{SoltauLiaoSak17}.
Such approaches are motivated by their simplicity: since these models directly
output graphemes, word-pieces~\cite{SchusterNakajima12}, or words, they do not
require expertly curated pronunuciation dictionaries; since they can be trained
to directly output normalized text, they do not require separate modules to map
recognized text from the spoken to the written domain.
In our recent work, we have shown that such approaches are comparable to
traditional state-of-the-art speech recognition
systems~\cite{RaoSakPrabhavalkar17, PrabhavalkarRaoSainathEtAl17}.

Most sequence-to-sequence models (e.g.,~\cite{ChanJaitlyLeEtAl15}) are typically
trained to optimize the cross-entropy (CE) loss function, which corresponds to
improving log-likelihood of the training data.
During inference, however, model performance is commonly measured using
task-specific criteria, not log-likelihood: e.g., word error rate (WER) for
ASR, or BLEU score~\cite{PapineniRoukosWardEtAl02} for machine translation.
Traditional ASR systems account for this mismatch through discriminative
sequence training of neural network acoustic models (AMs)~\cite{Kingsbury09,
VeselyGhoshalBurgetEtAl13} which fine-tunes a cross-entropy trained AM with
criteria such as state-level minimum Bayes risk (sMBR) which are more closely
related to word error rate.

In the context of sequence-to-sequence models, there have been a few previous
proposals to optimize task-specific losses.
In their seminal work, Graves and Jaitly~\cite{GravesJaitly14} minimize expected
WER of an RNN-T model by approximating the expectation with samples drawn from
the model.
This approach is similar to the edit-based minimum Bayes risk (EMBR) approach
proposed by Shannon, which was used for minimum expected WER training of
conventional ASR systems~\cite{Shannon17} and the recurrent neural
aligner~\cite{SakShannonRaoEtAl17}.
An alternative approach is based on reinforcement learning, where the label
output at each step can be viewed as an \emph{action}, so that the task of
learning consists of learning the \emph{optimal policy} (i.e., optimal output
label sequence) which results in the greatest expected reward (lowest expected
task-specific loss).
Ranzato et al.~\cite{RanzatoChopraAuliEtAl16} apply a variant of the REINFORCE
algorithm~\cite{Williams92} to optimize task-specific losses for summarization
and machine translation.
More recently Bahdanau et al.~\cite{BahdanauBrakelLoweEtAl17} use an
actor-critic approach, which was shown to improve BLEU scores for machine
translation.

In the present work, we consider techniques to optimize attention-based
sequence-to-sequence models in order to directly minimize WER.
Our proposed approach is similar to~\cite{GravesJaitly14, Shannon17} in that we
approximate the expected WER using hypotheses from the model.
We consider both the use of sampling-based approaches~\cite{GravesJaitly14,
Shannon17} as well as approximating the loss over N-best lists of recognition
hypotheses as is commonly done in ASR (e.g.,~\cite{Povey03}).
However, unlike Sak et al.~\cite{SakShannonRaoEtAl17} we find that the process
is more effective if we approximate the expectation using N-best hypotheses
decoded from the model using beam-search~\cite{SutskeverVinyalsLe14} rather than
sampling from the model (See section~\ref{sec:results1}).
We apply the proposed techniques on an English mobile voice-search task, to
optimize grapheme-based models, with uni- and bi-directional encoders, where we
find that we can improve WER by up to 8.2\% relative to a CE-trained baseline
model.
Minimum word error rate training allows us to train grapheme-based
sequence-to-sequence models which are comparable in performance to a strong
state-of-the-art context-dependent (CD) phoneme-based speech recognition
system~\cite{SeniorSakdeChaumontQuitryEtAl15}.

The organization of the rest of the paper is as follows.
We describe the particular attention-based model used in this work in
Section~\ref{sec:attention_models} and describe the proposed approach for
minimum WER training of attention models in Section~\ref{sec:embr}.
We describe our experimental setup and our results in
Sections~\ref{sec:experiments} and~\ref{sec:results}, respectively, before
concluding in Section~\ref{sec:conclusions}.

\section{Attention-Based Models}
\label{sec:attention_models}
We denote the set of speech utterances, suitably parameterized into feature
vectors as: $\x = (\x_1, \x_2, \cdots, \x_T)$, where $\x_i \in \R^{d}$, and the
corresponding ground-truth label sequence as: $\y^{*} = (y^{*}_0, y^{*}_1,
y^{*}_2, \cdots, y^{*}_{L+1})$, where $y^{*}_i \in \G$ (graphemes, in this
work).
We assume that the set of labels, $\G$, contains two special
labels, $\sos$ and $\eos$, which denote the start and the end of the sentence,
respectively, such that $y^{*}_0 = \sos$ and $y^{*}_{L+1}=\eos$.

\begin{figure}
  \centering
  \includegraphics[width=0.4\columnwidth]{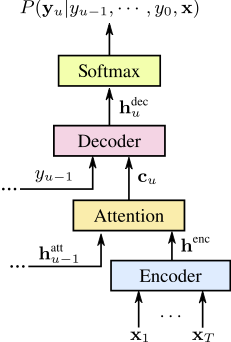}
  \caption{The attention-based model defines a probability distribution over the
  next label, conditioned on the history of previous predictions:
  $P(\y_u|y_{u-1}, \cdots, y_0, \x)$.}
  \label{fig:attention_model}
\end{figure}
An attention-based model~\cite{ChanJaitlyLeEtAl15} consists of three components:
an \emph{encoder network} which maps input acoustic vectors into a higher-level
representation, an \emph{attention model} which summarizes the output of the
encoder based on the current state of the decoder,  and a \emph{decoder network}
which models an output distribution over the next target conditioned on the
sequence of previous predictions: $P(\y_u|y^{*}_{u-1}, y^{*}_{u-2}, \cdots,
y^{*}_0, \x)$.
The model is depicted in Figure~\ref{fig:attention_model}.
The encoder network consists of a deep recurrent neural network which receives
as input the sequence of acoustic feature vectors, $\x$, and computes a
sequence of encoded features, $\h^\text{enc} = (\h^\text{enc}_1, \cdots,
\h^\text{enc}_T)$, and is analogous to an acoustic model in a traditional ASR
system.
The decoder network - which is analogous to the pronunication and language
modeling components in a traditional ASR system - consists of a deep recurrent
neural network which is augmented with an attention
mechanism~\cite{BahdanauChoBengio15}.
The decoder network predicts a single label at each step, conditioned on the
history of previous predictions.
At each prediction step, the attention mechanism summarizes the encoded features
based on the decoder state to compute a context vector, $\cv_{u}$, as described
in Section~\ref{sec:multi-headed-attention}.
The attention model thus corresponds to the component of a traditional ASR
system which learns the alignments between the input acoustics and the output
labels.
This context vector is input to the decoder along with the previous label,
$y^{*}_{u-1}$.
The final decoder layer produces a set of logits which are input to a softmax
layer which computes a distribution over the set of output labels: $P(\y_{u} |
y^{*}_{u-1}, \cdots, y^{*}_0=\sos)$.

\subsection{Multi-headed Attention}
\label{sec:multi-headed-attention}
The attention mechanism used in the present work differs from our previous
work~\cite{PrabhavalkarRaoSainathEtAl17} in two important ways: firstly, we
replace dot-product attention~\cite{ChanJaitlyLeEtAl15} with additive
attention~\cite{BahdanauChoBengio15} which we find to be more stable; secondly,
we use multiple, independent attention heads~\cite{VaswaniShazeerParmarEtAl17}
allowing the model to simultaneously attend to multiple locations in the
input utterance, which we find to significantly improve model
performance.
More specifically, we denote the recurrent hidden state of the decoder network
after predicting $u-1$ labels as $\h^\text{att}_{u-1}$.
The model employs $M$ independent attention heads, each of which computes
attention values, $\beta^{i}_{t, u} \in \R$, for $1 \leq i \leq M$, $1 \leq t
\leq T$:
\begin{equation}
\beta^{i}_{t, u} = \mathbf{u}^i \tanh(W^{i} \h^\text{att}_{u-1} + V^i\h^\text{enc}_t) \label{eq:additive-attention}
\end{equation}
The individual attention values are then transformed into soft attention weights
through a softmax operation, and used to compute a summary of the encoder
features, $\cv^{i}_u$:
\begin{equation}
\alpha^{i}_{t, u} = \frac{\exp(\beta^{i}_{t,
u})}{\sum_{s=1}^{T}\exp(\beta^{i}_{s, u})} \quad \quad
\cv^{i}_{u} = \sum_{t=1}^{T} \alpha^{i}_{t,u} Z^{i}\h^\text{enc}_t
\end{equation}
\noindent The matrices $V^{i}, W^{i}, \text{and } Z^{i}$ and the vector,
$\mathbf{u}^i$, are parameters of the model.
Finally, the overall context vector is computed by concatenating together the
individual summaries: $\cv_u = [\cv^{1}_u; \cv^{2}_u; \cdots; \cv^{M}_u]$.

\subsection{Training and Inference}
\label{sec:ce-loss}
Most attention-based models are trained by optimizing the cross-entropy (CE)
loss function, which maximizes the the log-likelihood of the training data:
\begin{equation}
\ll_\text{CE} = \sum_{(\x, \y^{*})} \sum_{u=1}^{L+1} -\log P(y^{*}_u | y^{*}_{u-1}, \cdots,
y^{*}_0=\sos, \x)
\end{equation}
\noindent where, we always input the ground-truth label sequence during
training (i.e., we do not use scheduled sampling~\cite{BengioVinyalsJaitlyEtAl15}).
Inference in the model is performed using a beam-search
algorithm~\cite{SutskeverVinyalsLe14}, where the models predictions are fed back
until the model outputs the $\eos$ symbol which indicates that inference is
complete.

\section{Minimum Word Error Rate Training of Attention-based Models}
\label{sec:embr}
In this section we described how an attention-based model can be trained to
minimize the expected number of word errors, and thus the word error rate.
We denote by $\wer(\y, \y^{*})$ the number of word errors in a hypothesis, $\y$,
relative to the ground-truth sequence, $\y^{*}$.
In order to minimize word error rates on test data, we consider as our loss
function, the \emph{expected number of word errors over the training set}:
\begin{equation}
\label{eq:embr}
\ll_\text{werr} (\x, \y^{*}) = \mathbb{E}[\wer(\y, \y^{*})] = \sum_{\y} P(\y|\x) \wer(\y, \y^{*})
\end{equation}
Computing the loss in~\eqref{eq:embr} exactly is intractable since it
involves a summation over all possible label sequences.
We therefore consider two possible approximations which ensure tractability:
\emph{approximating the expectation in~\eqref{eq:embr} with
samples}~\cite{SakShannonRaoEtAl17, Shannon17}, or restricting the summation to
an N-best list as is commonly done during sequence-training for
ASR~\cite{Povey03}.

\subsection{Approximation By Sampling}
We can approximate the expectation in~\eqref{eq:embr} using an empirical
average over samples drawn from the model~\cite{Shannon17}:
\begin{small}
\begin{equation}
\label{eq:embr-sampling}
\ll_\text{werr}(\x, \y^{*}) \approx \ll^\text{Sample}_\text{werr}(\x, \y^{*}) =
\frac{1}{N} \sum_{\y_i \sim P(\y|\x)} \wer(\y_i, \y^{*})
\end{equation}
\end{small}
\noindent where, $\y_i$ are N samples drawn from the model distribution.
Critically, the gradient of the expectation in~\eqref{eq:embr-sampling} can be
itself be expressed as an expectation, which allows it to be approximated using
samples~\cite{Shannon17}:
\begin{footnotesize}
\begin{align}
\nabla \ll^\text{Sample}_\text{werr} (\x, \y^{*})
  &= \sum_{\y} P(\y|\x) \left[\wer(\y, \y^{*}) - \mathbb{E}[\wer(\y,
  \y^{*})]\right] \nabla \log P(\y|\x) \nonumber \\
  &\approx \frac{1}{N} \sum_{\y_i \sim P(\y|\x)}
  [\wer(\y_i, \y^{*}) - \widehat{\wer}] \nabla \log P(\y|\x)  \label{eq:embr-grad-final}
\end{align}
\end{footnotesize}
\noindent where, we exploit the fact that
$\mathbb{E}[\nabla \log P(\y|\x)] = 0$, and $\widehat{\wer} = \frac{1}{N}
\sum_{i=1}^{N} \wer(\y_i, \y^{*})$ is the average number of word errors over the
samples.
Subtracting $\widehat{\wer}$, serves to reduce the variance of the gradient
estimates, and is important to stabilize training~\cite{Shannon17}.

\subsection{Approximation Using N-best Lists}
One of the potential disadvantages of the sampling-based approach is that a
large number of samples might be required in order to approximate the
expectation well.
However, since the probability mass is likely to be concentrated on the top-N
hypotheses, it is reasonable to approximate the loss function by restricting the
sum over just the top N hypotheses.
We note that this is typically done in traditional discriminative sequence
training approaches as well, where the summation is restricted to paths in a
lattice~\cite{Kingsbury09, VeselyGhoshalBurgetEtAl13}.

Denote by $\text{Beam}(\x, N) = \{\y_1, \cdots, \y_N\}$, the set of N-best
hypotheses computed using beam-search decoding~\cite{SutskeverVinyalsLe14} for
the input utterance $\x$, with a beam-size, $N$.
We can then approximate the loss function in~\eqref{eq:embr} by assuming that
the probability mass is concentrated on just the N-best hypotheses, as follows:
\begin{equation}
\label{eq:embr-nbest}
\ll^\text{N-best}_\text{werr} (\x, \y^{*}) = \sum_{\y_i \in \text{Beam}(\x, N)}
\widehat{P}(\y_i|\x) \left[\wer(\y_i, \y^{*}) - \widehat{W}\right] \nonumber
\end{equation}
\noindent Where, $\widehat{P}(\y_i|\x) = \frac{P(\y_i|\x)}{\sum_{\y_i \in
\text{Beam}(\x, N)} P(\y_i|\x)}$, represents the distribution re-normalized over
just the N-best hypotheses, and $\widehat{W}$ is the average number of word
errors over the N-best hypohtheses, which is applied as a form of variance
reduction, since it does not affect the gradient.

\subsection{Initialization and Training}
\label{sec:embr-init-train}
Based on the two schemes for approximating the expected word error rate, we can
define two possible loss functions:
\begin{equation}
\ll^\text{Sample} = \sum_{(\x, \y^{*})} \ll^\text{Sample}_\text{werr}(\x, \y^{*}) + \lambda \ll_\text{CE}
\end{equation}
\begin{equation}
\ll^\text{N-best} = \sum_{(\x, \y^{*})} \ll^\text{N-best}_\text{werr}(\x, \y^{*}) + \lambda \ll_\text{CE}
\end{equation}
In both cases, we interpolate with the CE loss function using a hyperparameter
$\lambda$ which we find is important to stabilize training (See
Section~\ref{sec:results}).
We note that interpolation with the CE loss function is similar to the
f-smoothing approach~\cite{SuLiYuEtAl13} in ASR.
Training the model directly to optimize $\ll^\text{Sample}$ or $\ll^\text{N-best}$
with random initialization is hard, since the model is not directly provided
with the ground-truth label sequence.
Therefore, we initialize the model with the parameters obtained after
CE training.

\section{Experimental Setup}
\label{sec:experiments}
The proposed approach is evaluated by conducting experiments on a mobile
voice-search task.
Models are trained on the same datasets as in our previous
works~\cite{PrabhavalkarRaoSainathEtAl17, PrabhavalkarSainathBoEtAl2017}.
The training set consists of $\sim$15M hand-transcribed anonymized utterances
extracted from Google voice-search traffic ($\sim$12,500 hours).
In order to improve robustness to noise, multi-style training data (MTR) are
constructed by artificially distorting training utterances with reverberation
and noise drawn from environmental recordings of daily events and from YouTube
using a room simulator, where the overall SNR ranges from 0-30dB with an average
SNR of 12dB~\cite{KimMisraChinEtAl17}.
Model hyperparameters are tuned on a development set of $\sim$12.9K utterances ($\sim$63K
words) and results are reported on a set of $\sim$14.8K utterances ($\sim$71.6K
words).

The acoustic input is parameterized into 80-dimensional log-Mel filterbank
features extracted over the 16kHz frequency range, computed with a 25ms window
and a 10ms frame shift.
Following~\cite{SakSeniorRaoEtAl15}, three consecutive frames are stacked
together, and every third stacked frame is presented as input to the encoder.
The same frontend is used for all models reported in this work.

Two attention-based models are trained in this work, differing only in the
structure of the encoder network: the first model (Uni-LAS) uses 5 layers of
1,400 uni-directional LSTM cells~\cite{HochreiterSchmidhuber97}, whereas the
second model (Bidi-LAS) uses 5 layers of 1,024 bi-directional LSTM
cells~\cite{SchusterPaliwal97} (i.e., 1,024 cells in the forward and backward
directions, for each layer).
The decoder network of both models consists of two layers of 1,024 LSTM cells in
each layer.
Both models use multi-headed attention as described in
Section~\ref{sec:multi-headed-attention} with $M=4$ attention heads.
Models are trained to output a probability distribution over grapheme symbols:
26 lower case alphabets \texttt{a-z}, the numerals \texttt{0-9},  punctuation
symbols \texttt{,'!} etc., and the special symbols $\sos$, $\eos$.
All models are trained using the Tensorflow
toolkit~\cite{AbadiAgarwalBarhamEtAl15}, with asynchronous stochastic gradient
descent (ASGD)~\cite{RechtReWrightEtAl11} using the Adam
optimizer~\cite{KingmaBa15}.

\section{Results}
\label{sec:results}
We investigate the impact of various hyperparameters, and the choice of
approximation scheme by conducting detailed
experiments on the uni-directional LAS model.
Results on the bi-directional LAS model, along with a comparison to a
traditional CD-phone based state-of-the-art system are deferred until
Section~\ref{sec:results2}.

\subsection{Comparison of loss functions: $\ll^\text{Sample}$ and
$\ll^\text{N-best}$}
\label{sec:results1}
\begin{figure*}
  \centering
  \begin{subfigure}[t]{0.33\textwidth}
    \includegraphics[width=\textwidth]{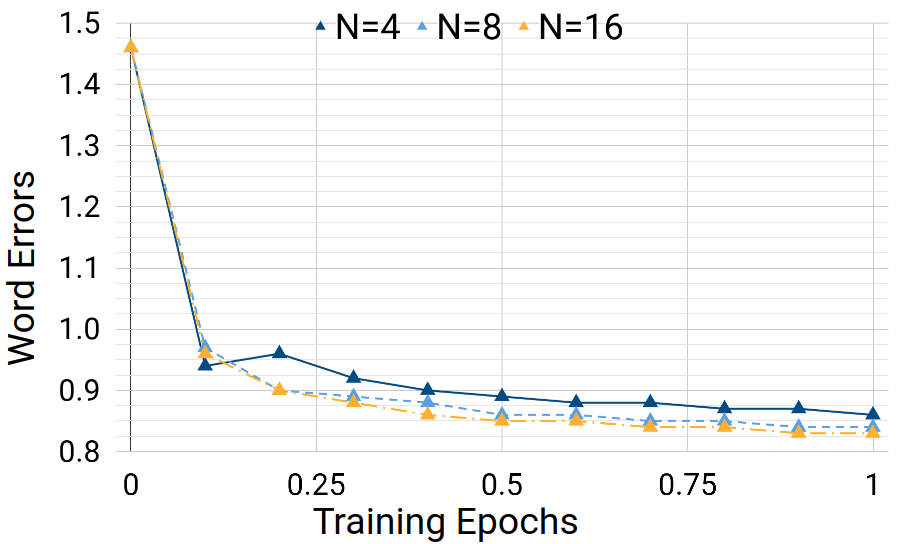}
    \caption{Expected number of word errors on held-out set computed using~\eqref{eq:embr} when
    	     optimizing $\ll^\text{Sample}$ as number of samples, $N$, varies.}
    \label{fig:nwerr_samp}
    \end{subfigure}
    ~
    \begin{subfigure}[t]{0.31\textwidth}
    \includegraphics[width=\textwidth]{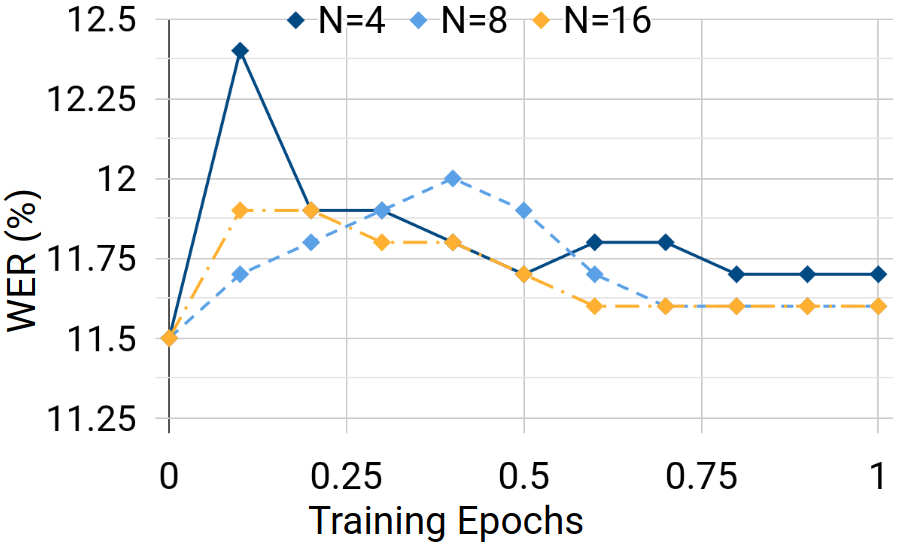}
    \caption{Word error rates on held-out set when optimizing $\ll^\text{Sample}$ as a function
    	     of the number of samples, $N$.}
    \label{fig:wer_samp}
    \end{subfigure}
    ~
    \begin{subfigure}[t]{0.31\textwidth}
    \includegraphics[width=\textwidth]{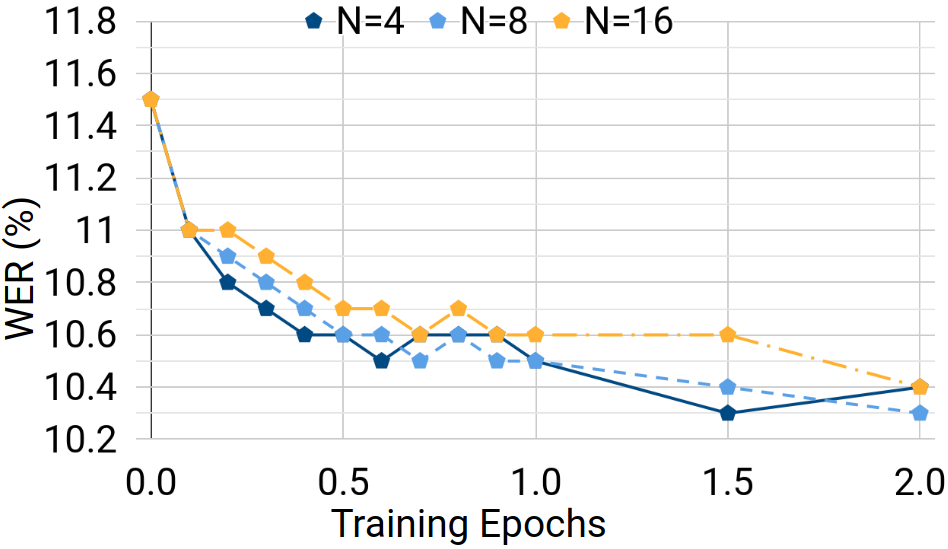}
    \caption{Word error rates on held-out set when optimizing $\ll^\text{N-best}$ as a function of the depth of the
	     N-best list, $N$.}
    \label{fig:wer_nbest}
    \end{subfigure}
    \caption{Metrics computed on held-out portion of the training set when
    	     optimizing loss functions $\ll^\text{Sample}$ and
	     $\ll^\text{N-best}$,
	     described in Section~\ref{sec:embr-init-train}.}
    \label{fig:metrics}
\end{figure*}
Our first set of experiments evaluate the effectiveness of approximating the
expected number of word errors using samples (i.e., optimizing $\ll^\text{Sample}$)
versus the approximation using N-best lists (i.e., optimizing
$\ll^\text{N-best}$), as described in Section~\ref{sec:embr-init-train}.
Our observations are illustrated in Figure~\ref{fig:metrics}, where we plot
various metrics on a held-out portion of the training data.

As can be seen in Figure~\ref{fig:nwerr_samp}, optimizing the sample-based
approximation, $\ll^\text{Sample}$, reduces the expected number of word errors by
$\sim$50\% after training, with performance appearing to improve as the number
of samples, $N$, used in the approximation increases.
Unlike~\cite{SakShannonRaoEtAl17}, however, as can be seen in
Figure~\ref{fig:wer_samp}, the WER for the top-hypothesis computed using beam
search does not improve, but instead degrades as a result of training.
We hypothesize that this is a result of the mis-match between the beam-search
decoding procedure, which focuses on the head of the distribution during each
next-label prediction, and the sampling procedure which also considers
lower-probability paths~\cite{RanzatoChopraAuliEtAl16}.

As illustrated in Figure~\ref{fig:wer_nbest}, optimizing $\ll^\text{N-best}$
(i.e., using the N-best list-based approximation) significantly improves WER by
about 10.4\% on the held-out portion of the training set.
Further, performance seems to be similar even when just the top four hypotheses
are considered during the optimization.

\begin{figure}
  \centering
  \includegraphics[width=0.75\columnwidth]{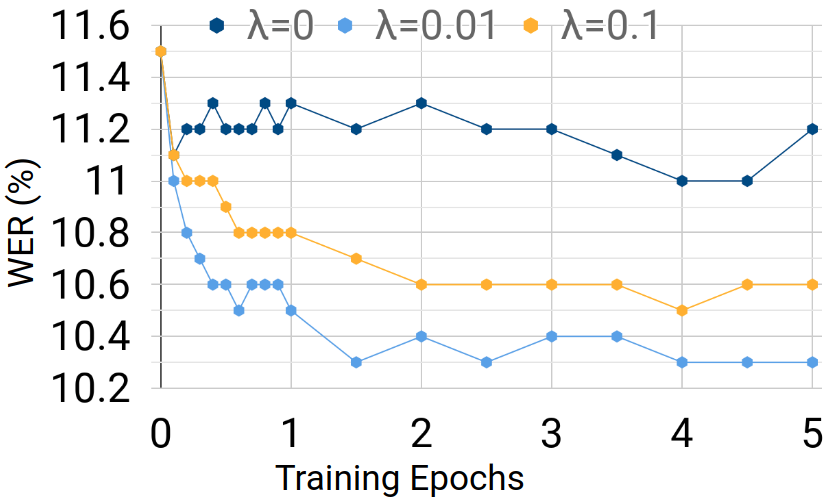}
    \caption{Word error rates on held-out portion of training set when
    optimizing $\ll^\text{N-best}$, as a function of the CE-loss interpolation
    weight $\lambda$, when using $N=4$ hypotheses in the N-best list.}
    \label{fig:wer_lambda_nbest}
\end{figure}
As a final note, we find that it is important to also interpolate with CE loss
function during optimization (i.e., setting $\lambda > 0$).
This is illustrated for the case where we optimize $\ll^\text{N-best}$ using
$N=4$ hypotheses in the N-best list in Figure~\ref{fig:wer_lambda_nbest}.

\subsection{Improvements from Minimum WER Training for LAS Models}
\label{sec:results2}
\begin{table}
  \centering
  \begin{tabular}{|c||c||c|}
    \hline
    System & WER(\%) & Rescored WER(\%) \\
    \hline
    \hline
    Bi-LAS & 7.2  & 6.6 \\
    +MWER ($\ll^\text{N-best})$ & 6.9 & 6.2 \\
    \hline
    \hline
    Uni-LAS & 8.1 & 7.3 \\
    +MWER ($\ll^\text{N-best}) $ & 7.5 & 6.7 \\
    \hline
    \hline
    CD-phone (CE + sMBR) & 7.5 & 6.7 \\
    \hline
  \end{tabular}
  \caption{WERs on the test set after minimum WER training for uni- and
	   bi-directional LAS models. The proposed procedure improves WER by up
	   to 8.2\% relative to the CE-trained baseline system.}
  \label{tbl:results}
\end{table}
We present results after expected minimum WER training (MWER) of the uni- and
bi-directional LAS models described in Section~\ref{sec:experiments} in
Table~\ref{tbl:results}, where we set $N=4$ and $\lambda=0.01$.
We report results after directly decoding the models to produce grapheme
sequences using a beam-search decoding with 8 beams (column 2) as well as after
rescoring the 8-best list using a very large 5-gram language model (column 3).
For comparison, we also report results using a traditional state-of-the-art low
frame rate (LFR)~\cite{PundakSainath16} CD-phone based system, which uses an
acoustic model composed of four layers of 1,024 uni-directional LSTM cells,
followed by one layer of 768 uni-directional cells.
The model is first trained to optimize the CE loss function, followed by
discriminative sequence training to optimize the state-level minimum Bayes risk
(sMBR) criterion~\cite{Kingsbury09}.
The model is decoded using a pruned, first-pass, 5-gram language model, which
uses a vocabulary of millions of words, as well as an expert-curated
pronunciation dictionary.
As before, we report results both before and after second-pass lattice
rescoring.

As can be seen in Table~\ref{tbl:results}, when decoded without second-pass rescoring (i.e.,
end-to-end training), MWER training improves performance of the uni- and
bi-directional LAS systems by 7.4\% and 4.2\% respectively.
The gains after MWER training are even larger after second-pass rescoring,
improving the baseline uni- and bi-directional LAS systems by 8.2\% and 6.1\%,
respectively.
Finally, we note that after MWER training the grapheme-based uni-directional LAS
system matches the performance of a state-of-the-art traditional
CD-phoneme-based ASR system.

\section{Conclusions}
\label{sec:conclusions}
We described a technique for training sequence-to-sequence systems to optmize
the expected test error rate, which was applied to attention-based systems.
Unlike~\cite{SakShannonRaoEtAl17}, we find that sampling-based approximations
are not as effective as approximations based on using N-best decoded hypotheses.
Overall, we find that the proposed approach allows us to improve WER by up to
8.2\% relative.
We find that the proposed techniques allow us to train grapheme-based
sequence-to-sequence models which match performance with a traditional
CD-phone-based state-of-the-art system on a voice-search task, which when viewed
jointly with our previous works~\cite{PrabhavalkarRaoSainathEtAl17,
RaoSakPrabhavalkar17} adds further evidence to the effectiveness of
sequence-to-sequence modeling approaches.


\vfill\pagebreak

\bibliographystyle{IEEEbib}
\bibliography{main}

\end{document}